\documentclass[a4paper,conference]{IEEEtran}

\usepackage[T1]{fontenc}
\usepackage{graphicx}
  \graphicspath{{figures/}}
\usepackage{url}
\usepackage{tabularx}
\usepackage{booktabs}
\usepackage{mathtools}
\usepackage{amsmath}
  \allowdisplaybreaks
\usepackage{amsfonts}
\usepackage{amssymb}
\usepackage{bm}
\usepackage[dvipsnames]{xcolor}
\usepackage{soul}
\usepackage{svg}
\usepackage{array}
\usepackage{hyperref}
\usepackage{float}
\usepackage{stfloats}
\usepackage{listings} % For formatting code
\usepackage{xcolor} % For coloring code
\usepackage{graphicx}
\usepackage[font=small]{caption}
\usepackage{multirow} % For multi-row spanning

\makeatletter
\def\footnoterule{\relax%
  \kern-5pt
  \hrule\@width.4\columnwidth
  \kern4.6pt}
\makeatother

\makeatletter
\def\subsubsection{%
  \@startsection
    {subsubsection}                 % type
    {3}                             % level
    {\parindent}                    % indent
    {0.5ex plus 1.5ex minus 1.5ex}  % beforeskip {0ex plus 0.1ex minus 0.1ex}
    {0.7ex plus .5ex minus 0ex}     % afterskip {0ex}
    {\normalfont\normalsize\itshape}% style
}
\makeatother

\usepackage[scaled=0.8]{FiraMono}

\definecolor{gray90}{gray}{0.90}

\renewcommand{\texttt}[1]{\colorbox{gray90}{{{\ttfamily #1}}}}

\title{
Automatic Control With Human-Like Reasoning:\\ 
\huge Exploring Language Model Embodied Air Traffic Agents
}

\author{
\IEEEauthorblockN{Justas Andriuškevičius, Junzi Sun\IEEEauthorrefmark{1}}
\IEEEauthorblockA{Faculty of Aerospace Engineering, Delft University of Technology\\
Delft, the Netherlands\\
Corresponding Email: \IEEEauthorrefmark{1}j.sun-1@tudelft.nl
}
}

\begin{document}

\maketitle

\begin{abstract}
Recent developments in language models have created new opportunities in air traffic control studies. The current focus is primarily on text and language-based use cases. However, these language models may offer a higher potential impact in the air traffic control domain, thanks to their ability to interact with air traffic environments in an embodied agent form. They also provide a language-like reasoning capability to explain their decisions, which has been a significant roadblock for the implementation of automatic air traffic control.

This paper investigates the application of a language model-based agent with function-calling and learning capabilities to resolve air traffic conflicts without human intervention. The main components of this research are foundational large language models, tools that allow the agent to interact with the simulator, and a new concept, the experience library. An innovative part of this research, the experience library, is a vector database that stores synthesized knowledge that agents have learned from interactions with the simulations and language models.

To evaluate the performance of our language model-based agent, both open-source and closed-source models were tested. The results of our study reveal significant differences in performance across various configurations of the language model-based agents. The best-performing configuration was able to solve almost all 120 but one imminent conflict scenarios, including up to four aircraft at the same time. Most importantly, the agents are able to provide human-level text explanations on traffic situations and conflict resolution strategies.
\end{abstract}

{\small\textbf{\textit{keywords} --} Air traffic control, self-learning agents, large language models, experience library, function-calling}

\section{Introduction}
\label{sec:Introduction}
Air traffic management is a system that is critical for ensuring global airspace safety and operational efficiency. As air traffic volumes increase, so does the complexity of managing numerous flights and workloads for operators simultaneously \cite{SKALTSAS201346}, which raises the risk of incidents due to operational misunderstandings. These factors have historically contributed significantly to aviation accidents.

One of the main developments in air traffic management is the introduction of artificial intelligence in air traffic control to reduce the workload of air traffic controllers. The SESAR AISA project \cite{aisa2022} was an early attempt to incorporate AI into air traffic management by creating a system for artificial situational awareness through the use of knowledge graphs and machine learning for traffic prediction. The SESAR TAPAS project \cite{zou2023investigating} represented an advancement in the ATM field, targeting explainability. The project tested explainable AI and visual analytics in human-operated simulations that tried to make AI's decision-making processes accessible to controllers. Similarly, another SESAR project, ARTIMATION \cite{Hurter2022}, also aims at producing a transparent AI through visualization.

Overall, the human-in-the-loop simulations revealed a gap between artificial and human situational awareness, highlighting room for improvement in AI’s complex decision-making processes. This gap requires AI to offer more nuanced and human-like reasoning capabilities in air traffic management.

Since 2023, researchers have experimented with the integration of large language models (LLM) into air traffic management. Large language models are advanced AI systems capable of understanding and generating human-like text. Their proficiency in real-time decision-making has the potential to improve operational efficiency and automate labor-intensive tasks. Most of the data used to train leading-edge large language models, such as the latest Common Crawl dataset \cite{commoncrawl}, which comprises over 250 billion web pages, sources information from publicly accessible internet sites. 

This extensive training equips large language models with a broad understanding of air traffic management standards, including guidelines from the International Civil Aviation Organisation, Federal Aviation Administration regulations, and other global and local aviation protocols. Consequently, large language models can interpret these contents effectively.

Several recent studies have explored use cases for aviation applications. For example, \cite{abdulhak2024chatatc} employs language models to understand ground delay program text data. \cite{wang2024aviationgpt} fine-tunes the open-source language models to better understand the aviation context. A recent study \cite{jarry2024effectiveness} uses a language model for text classification and clustering based on air traffic flow management regulations and weather reports.

However, these use cases are primarily focused on natural language processing; they have not utilized the full potential of language models in managing air traffic operations nor looked into how AI can provide human-like reasoning. 

A new concept, the language model embodied agent, Voyager \cite{voyager}, was introduced last year, which represents an innovative step in leveraging the language model's reasoning capability. It is designed for open interactions within the Minecraft game environment, where Voyager agents can explore the virtual world autonomously and, most importantly, acquire skills by experience and then apply skills.

In a similar context, we also hypothesize that large language models may act as intelligent assistants for air traffic control operators, helping to manage routine tasks. More critically, these agents can play a decisive role in conflict resolution strategies—identifying potential conflicts and suggesting optimal maneuvering strategies. By leveraging the function-calling capability of the language model, they can also interact with the air traffic simulator and start learning air traffic control experiences like a new air traffic controller in training.

% The main advantage of large language models over other artificial intelligence (AI) systems, such as reinforcement learning algorithms, decision trees, and neural networks, is their interpretability. Unlike these other AI forms, which often function as black-box systems with unclear processes, large language models can display their reasoning and decision-making processes step-by-step. This transparency allows for deeper insights into how conclusions are drawn.

This paper explores a novel application of the large language model embodied agents in air traffic control. Our agent is able to interact with air traffic scenarios, monitor traffic, build up experiences, and resolve conflicts, all the while providing reasons for its behavior like an air traffic controller. The study assesses how effectively large language model agents can resolve air traffic conflicts and discusses in detail the limitations and potential for adopting our approach with human-like reasoning capabilities to assist air traffic controllers.

% The structure of the paper is as follows. Section \label{sec:method} explains the main concept of language model embodied air traffic control agents and how they gain experiences and provide commands based on math and reasons. Section \ref{sec:experiments} explains the details of our experiments in using different types of agent architectures to solve aircraft conflicts. We will also provide in-depth discussion in Section \ref{sec:discussions} before concluding our work in Section \ref{sec:conclusion}.

\section{Methodology}  \label{sec:method}

In this section, we discuss our efforts to develop two different large language model embodied agent frameworks, which are capable of interacting with the BlueSky simulator~\cite{hoekstra2016bluesky}, monitoring and interpreting traffic situations automatically, and producing instructions to solve air traffic conflicts autonomously and in real-time. 

\subsection{Large language model embodied Agent}
A language model predicts the next word in a sequence by analyzing the preceding words. Increasing the complexity of models, like large transformer models, leads to awareness of the extremely long context in text. The text includes programming language and software code. By providing a proper application programming interface, these models can be integrated with various tools and virtual or real environments, transforming them into embodied agents. An embodied agent can either utilize specific tools, such as Python functions with arguments or operate independently to generate responses. 

In our research, we designed such agents that can interact with the air traffic control interface, for example, the BlueSky simulator. By providing the proper objective in a text (called \emph{prompt}), the agent is set to solve conflict scenarios.

\begin{figure}
    \centering
    \includegraphics[width=\linewidth]{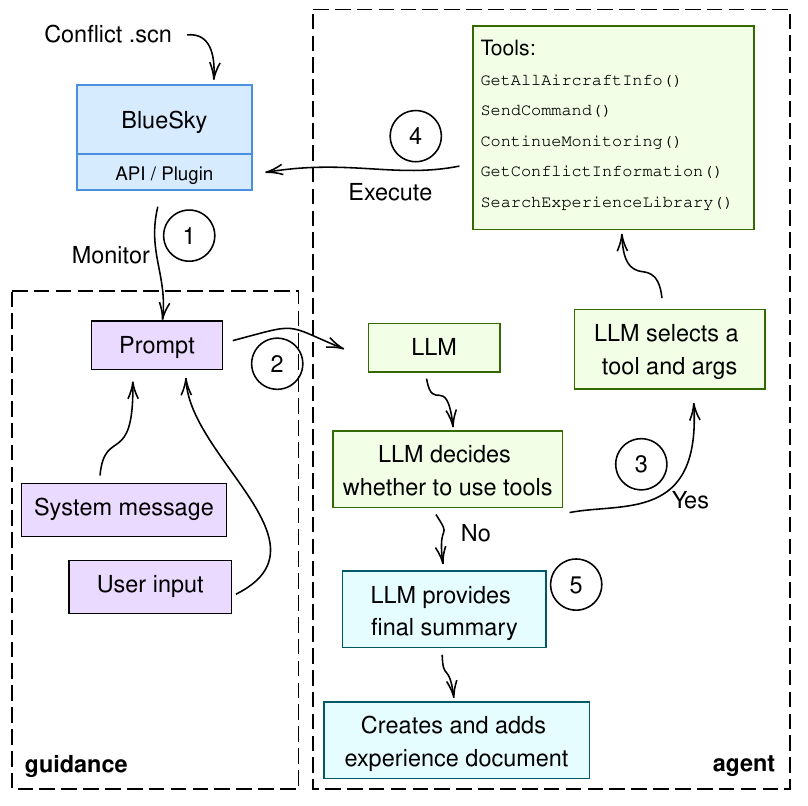}
    \caption{The language model embodied single agent setup}
    \label{fig:single_agent_overview}
\end{figure}

\autoref{fig:single_agent_overview} shows the overview of the process, beginning with the construction of a prompt that integrates the system prompt, user prompt, and tools descriptions. The large language model then evaluates whether a tool is needed for the task at hand. If a tool is required, the agent executes the selected tool with the specified arguments. For example, to change an aircraft's altitude, the agent would use a \texttt{SendCommand()} tool with a generated altitude command. This command is sent to the simulator, and the output from the simulator is then integrated back into the prompt for further processing. This cycle repeats until the large language model determines that no additional tools are needed. 

In the end, the agent can also provide a summary of the situation and reasons for the conflict-solving strategies. An experience document (\autoref{sub:experience_library}) is then created and subsequently uploaded to an experience library, which can be retried to further enhance the agent's knowledge base and capabilities for more complex tasks.

\begin{figure*}[ht!]
    \centering
    \includegraphics[width=0.95\linewidth]{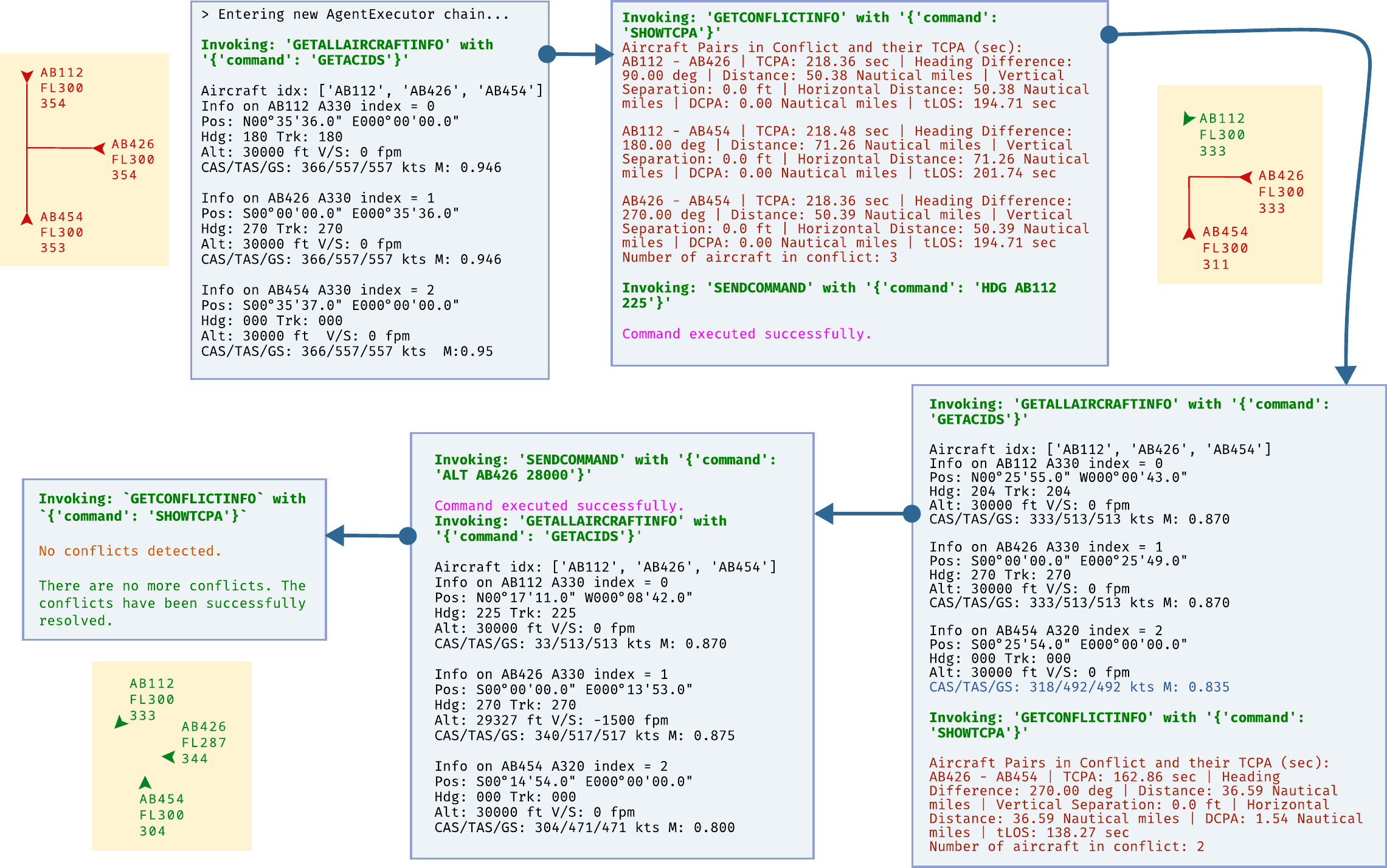}
    \caption{Single Agent solving 3 aircraft conflicts without experience library. The LLM embodied agent automatically decides when and what commands (in green text) are to be invoked at all stages.}
    \label{fig:llm_single_agent_3ac_wo_ex}
\end{figure*}

To demonstrate this process, \autoref{fig:llm_single_agent_3ac_wo_ex} presents a scenario where a single agent effectively resolves a converging three-aircraft conflict. The resolution process begins with the agent querying all relevant aircraft data through the \texttt{GetAllAircraftInfo()} tool. The agent automatically assesses the conflict dynamics between each pair of aircraft using \texttt{GetConflictInfo()}. Based on the results, the agent then strategically issues a heading change to aircraft AB112, directing it to alter its course to 225 degrees. This directive is executed via the \texttt{SendCommand()} tool, utilizing the command \texttt{HDG AB112 225}. 

After this initial conflict mitigation, the agent re-evaluates the aircraft and conflict information. It then proceeds to issue another command - this time decreasing the altitude of aircraft AB426 by 2000 feet, further solving the remaining conflict. After re-assessing the situation and confirming the resolution of all potential conflicts, the agent concludes its task, having successfully ensured a safe outcome. 

\begin{figure}[ht!]
    \centering
    \includegraphics[width=\columnwidth]{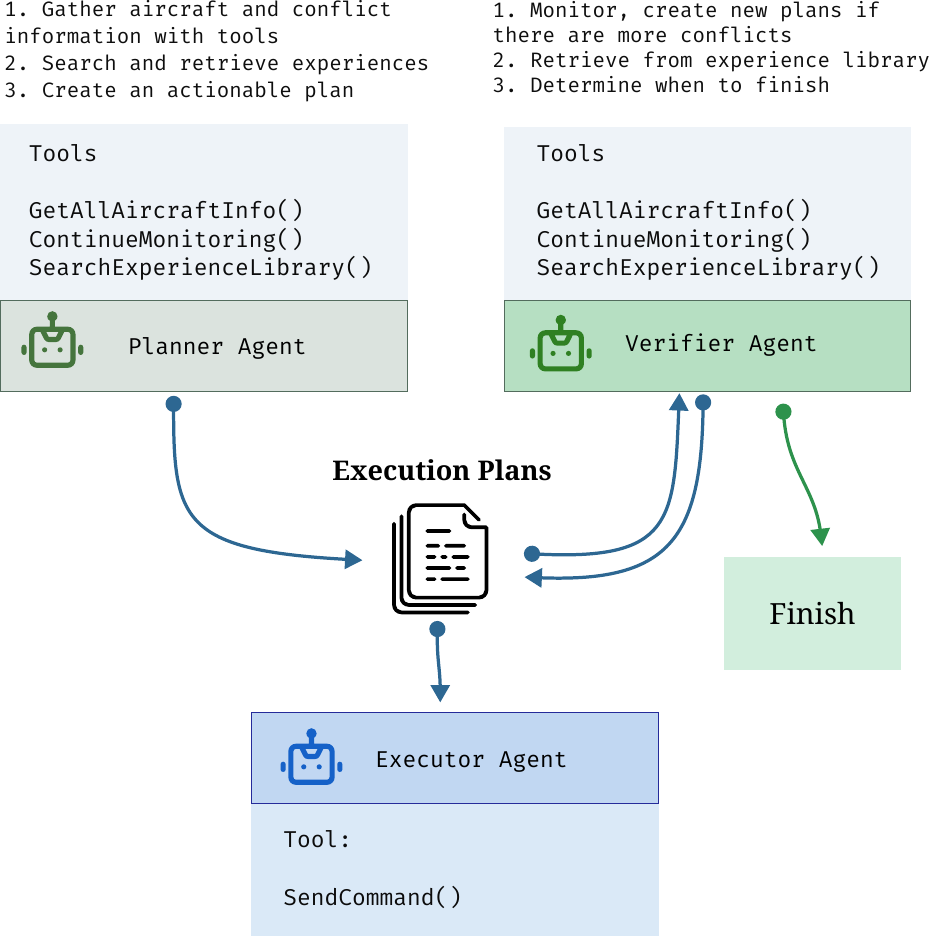}
    \caption{The structure of the multiple language model embodied agent, containing planner, verifier, and executor agents.}
    \label{fig:multi_agent_overview}
\end{figure}

We have also developed a multi-agent system capable of handling an unrestricted number of LLM embodied agents and facilitating increasingly complex challenges. This system is illustrated in \autoref{fig:multi_agent_overview}. 

In this multi-agent system, we designed three types of agents: the planner, the executor, and the verifier. The planner agent is responsible for generating a conflict resolution plan. It begins this process by monitoring the airspace and analyzing detected conflicts. Once a plan is formulated, it is passed onto the executor agent. The sole function of this agent is to issue appropriate commands to BlueSky. 

After the execution of the plan, the verifier agent plays a critical role in ensuring the efficacy of the conflict resolution. This agent continues to monitor the airspace to confirm whether any conflicts remain unresolved. If conflicts persist, the verifier agent devises a new resolution plan, which is once again forwarded to the executor agent for implementation. Conversely, if no further conflicts are detected, the conflict-solving task is concluded. Additionally, when the experience library is activated, both the planner and verifier agents can search in this library to retrieve insights from previously encountered conflicts.

\subsection{Prompt}
The prompt serves as a critical link between the objectives, agent actions, and the underlying language model. We have designed a prompt template to ensure the clarity and relevance of the information processed by the large language model, containing four different components:

\begin{lstlisting}
system_prompt:    pre-crafted text on role and objectives
user_input:       instructions from human
chat_history:     memories about llm inputs and outputs
agent_scratchpad: memories about environment interactions 
\end{lstlisting}

\textbf{System Prompt:} This component is crafted to provide both context and explicit instructions to the agent. Each agent receives tailored directives specific to their role. For instance, the planner agent is instructed to gather aircraft information, monitor airspace, and provide an actionable plan according to the separation requirements that would also avoid introducing new conflicts.

\textbf{User Input:} A brief on tasks and preferences that can enhance agent performance. For instance, the planner agent may be asked to check for conflicts and create a plan based on preferences, such as changing heading, altitude, or both. These instructions are less detailed than the system prompt.

\textbf{Chat History:} Acting as a memory block, which stores the inputs and the output with the language model, it maintains a continuous record of interactions.

\textbf{Agent Scratchpad:} This component memorizes descriptions of the tools used, logs all intermediate steps, and records results from the tools. It is vital for tracking the agent's operational processes and the adaptations made during task execution.

% \begin{tcolorbox}[sharp corners, boxrule=0.5pt, colback=white, colframe=black!75!white, title=System Prompt]

% You are an air traffic control assistant. Your goal is to solve aircraft conflict according to the following requirements: either vertical separation of 2000 ft or horizontal separation of 5 nautical miles between aircraft.
% \\\\
% Command to change aircraft altitude is:\\ ALT AIRCRAFT\_CALL\_SIGN ALTITUDE. Command to change aircraft heading is:\\ HDG AIRCRAFT\_CALL\_SIGN HEADING. ALTITUDE is in feet and HEADING is in degrees. The aircraft call sign is a unique identifier for each aircraft.
% \\\\
% You need to send commands in order to solve the conflicts. You must solve the conflicts till there are no more conflicts left.
% \\\\
% Use QueryConflicts tool to look for similar conflicts in the database that where already solved, it can help you to solve the current conflicts.
% \end{tcolorbox}

\subsection{Tools}
\label{sub:tools}
Our system integrates several specialized tools (functions in Python programming language) to facilitate interactions between the large language model and the BlueSky simulator. These tools are crucial for the effective execution of tasks and data retrieval:

\begin{itemize}
\item \texttt{GetAllAircraftInfo()}: This tool sends a command to BlueSky and retrieves a comprehensive list of aircraft, detailing their position, heading, track, altitude, vertical speed, calibrated, true airspeed, and ground speed, as well as Mach number.
\item \texttt{GetConflictInfo()}: This tool sends a command to BlueSky and retrieves information about aircraft pairs in conflict. It provides details such as Time to Closest Point of Approach (TCPA), heading differences, separation distances (total, vertical, and horizontal), distance to Closest Point of Approach distance (DCPA), time to Loss of Separation (tLOS), and altitude information.
\item \texttt{ContinueMonitoring(duration)}: This tool commands BlueSky to retrieve changes in conflict status over a specified duration, enabling ongoing monitoring of the airspace.
\item \texttt{SendCommand(command)}: This tool sends a traffic command to BlueSky and retrieves the resulting output from the simulator, allowing for dynamic interaction with the simulation environment.
\item \texttt{SearchExperienceLibrary(args)}: This tool queries the experience library and returns the most relevant experience document based on different arguments, including conflict description, number of aircraft involved, and the formation of the conflict.
\end{itemize}

It is important to emphasize that the large language model decides when to utilize a tool, and it is also responsible for generating proper functional arguments that enable precise and context-appropriate responses. This function-calling capability enhances the agent's ability to interact with and manipulate the environment effectively and freely. 

In principle, it is also possible for the agent to write its own tools, considering that sufficiently large language models are also capable of code generation. However, this was not tested in our experiments.

\subsection{Experience Library}
\label{sub:experience_library}

\begin{figure*}[ht!]
    \centering
    \includegraphics[width=0.95\linewidth]{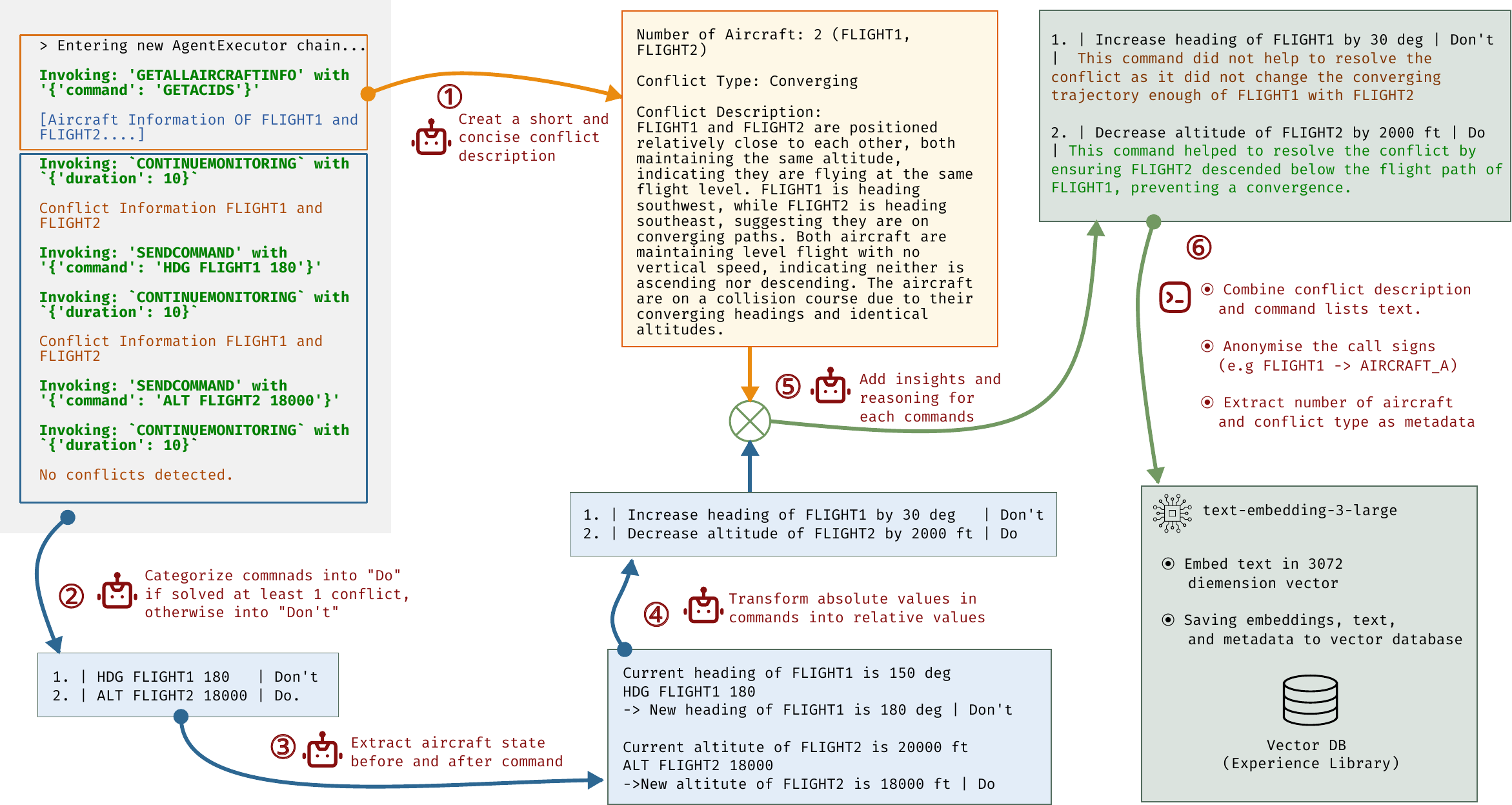}
    \caption{Creating the experience document from the operation logs of the agent}
    \label{fig:exp_doc_creation}
\end{figure*}

The Experience Library is a crucial component that enables our LLM embodied agent to recall stored memories about past conflict solution experiences. We use an open vector database, \texttt{Chroma} \cite{chroma}, to store and retrieve past conflict resolutions effectively. A vector database encodes text (a.k.a. tokens) into numerical vectors, which can be compared based on similarities. The agent can create and search the experience library on its own.

\subsubsection{Creation of Experience Documents} 

After an LLM agent resolves a conflict, it processes the entire conflict resolution log to create an \emph{experience document}. A concise conflict description is generated with the language model based on the initial states of the aircraft and the conflict information. It then categorizes the executed commands into whether they are helpful or not helpful. 

Commands that have eliminated at least one conflict pair are deemed helpful, while others are not. The absolute values (like altitudes and headings) of these commands are converted into relative values. The conflict description and the categorized list of commands are then combined. Finally, the language model enhances the document by adding insights and reasoning for each command tailored to the specific conflict description. Additionally, aircraft callsigns are anonymized in the final steps of creating the experience document. This ensures that when an agent retrieves the document later, it won't be confused by the presence of the same callsigns in both the current conflict and the experience document. 

The conflict description is encoded into a 3072-dimensional vector embedding using the \texttt{text-embedding-3-large} model from OpenAI\footnote{Many other embedding models can be used, for example, \url{https://huggingface.co/models?other=text-embeddings-inference}}. The embeddings of the experience, along with text and metadata on conflict type and the number of aircraft, are then uploaded to the vector database. 

The entire experience generation process is illustrated in \autoref{fig:exp_doc_creation}. It is worth noting that we only need to encode the conflict description. This is because when an agent searches the experience library, it can describe the current conflict. Matching conflict descriptions directly yields higher similarity and the most relevant results than when comparing the full document with commands, suggestions, and insights.

\subsubsection{Experience Library Search} 

\begin{figure}[ht!]
    \centering
    \includegraphics[width=0.95\columnwidth]{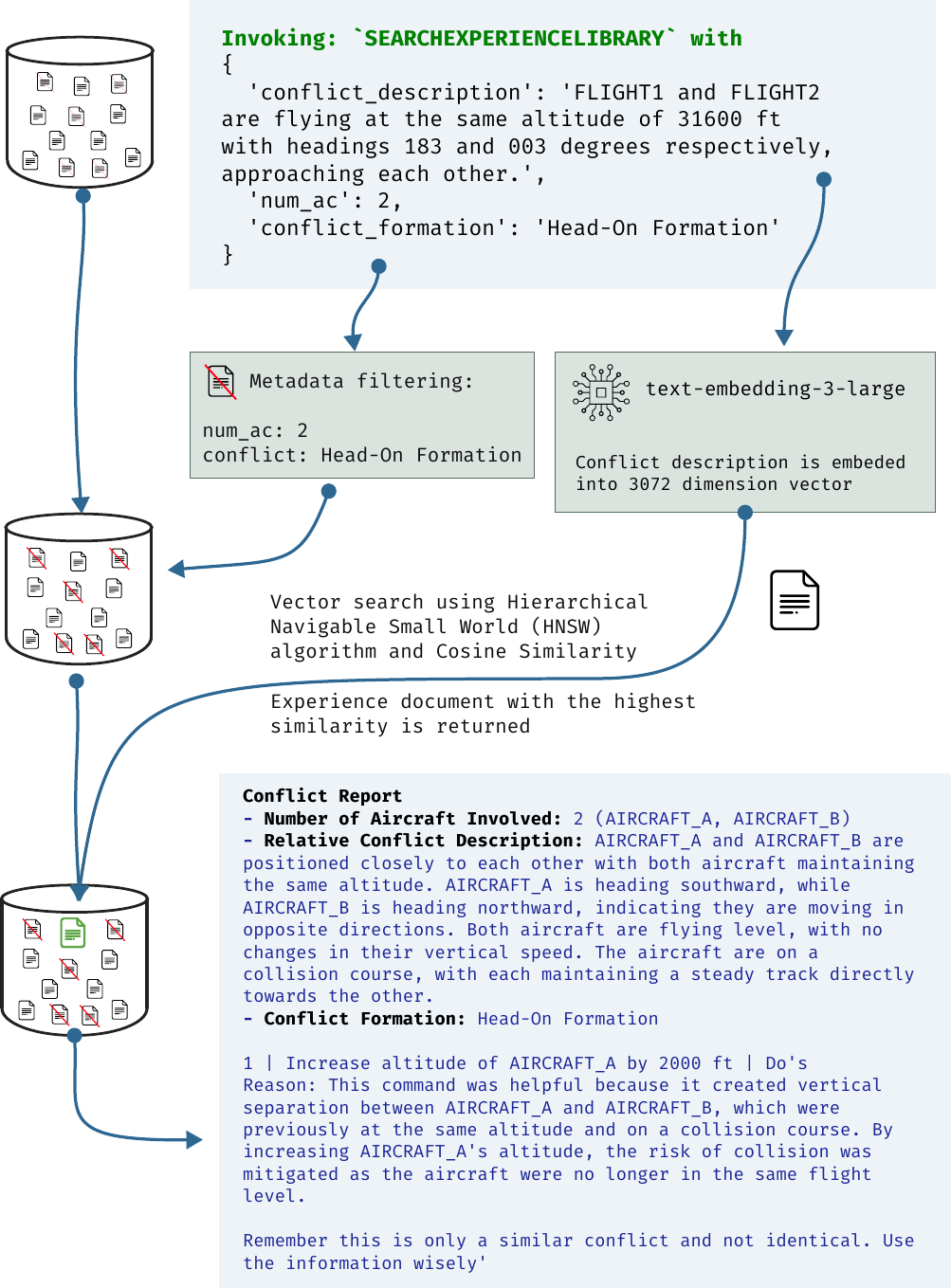}
    \caption{Filtering and searching in the experience library based on experience embeddings}
    \label{fig:exp_lib_search}
\end{figure}

When an agent wants to retrieve the closest memory from past experiences before solving the conflicts, it invokes the \texttt{SearchExperienceLibrary()} tool (shown in \autoref{fig:exp_lib_search}). The agent first generates a concise description of the current conflict, including the number of aircraft involved and the type of conflict. The initial metadata filtering reduces the search space in terms of aircraft formation and number of aircraft. The conflict description is also encoded as a 3072-dimensional vector with the embedding model. 

The search process employs the Hierarchical Navigable Small World (HNSW) \cite{HNSW} algorithm alongside Cosine Similarity to perform the vector search. The system returns the experience document that exhibits the highest textual similarity.

\section{Experiments and results} \label{sec:experiments}

In this section, we describe the experimental setup and the results of various agent configurations under many simulated conflict scenarios. The performances of different agent models with and without access to the Experience Library are explored. Our experiments are structured to assess the effectiveness of addressing a range of increasingly complex scenarios. 

\subsection{Initial tests}

An initial experiment was conducted with a small dataset containing 12 conflict scenarios, which included four types of conflicts (head-on, parallel, t-formation, converging) with three different aircraft numbers each (2, 3, and 4 aircraft). We first tested a single agent configuration without experience library with the following models: \texttt{Llama3:7B}, \texttt{Llama3:70B}, \texttt{Mixtral
8x7b}, \texttt{gemma2:9b-it} and \texttt{GPT-4o}. 

We also evaluated different range of temperatures: \texttt{0.0}, \texttt{0.3}, \texttt{0.6}, \texttt{0.9}, and \texttt{1.2}. The temperature determines how conservative a language model predicts the next token. The higher the temperature, the more \emph{creative} the agents become. 

This initial experiment was designed to identify the most promising models based on a limited set of scenarios. The tests narrow down the number of models to focus on for later more extensive testing. We score the effectiveness of the setting based on the criteria in \autoref{tab:scoring_system}.

\begin{table}[ht!]
\centering
\caption{Scores for evaluating conflict resolution}
\label{tab:scoring_system}
\begin{tabularx}{\columnwidth}{lX}
\toprule
\textbf{Score} & \textbf{Outcome Description} \\
\midrule
1 & Conflict is solved (successful) \\
0 & Conflict results in loss of separation (unsuccessful) \\
-1 & Conflict results in near miss or collision (unsuccessful) \\
\bottomrule
\end{tabularx}
\end{table}

Based on these tests, the \texttt{Llama3:70B} (open-source) and \texttt{GPT-4o} (commercial) models exhibited the best performance success rates. And the temperature of \texttt{0.3} provides the most stable results. 

%% results table can be included for the final thesis / journal paper

\subsection{Generating large conflict scenarios}

To assess the performance of the \texttt{Llama3:70B} and \texttt{GPT-4o} models in solving air traffic conflicts, we generated a dataset comprised of 120 distinct conflict scenarios for BlueSky.

The dataset contains 40 scenarios, each with two, three, or four aircraft conflicts. The conflicts are categorized into four primary types: 1) head-on, where aircraft are on a direct collision course; 2) T-formation, which involves perpendicular flight paths; 3) parallel, where aircraft fly close parallel courses; and 4) converging, where multiple aircraft are on intersecting paths heading towards the same point. There are 30 conflict scenarios in these four types.

\begin{figure}[ht!]
    \centering
    \includegraphics[width=\columnwidth]{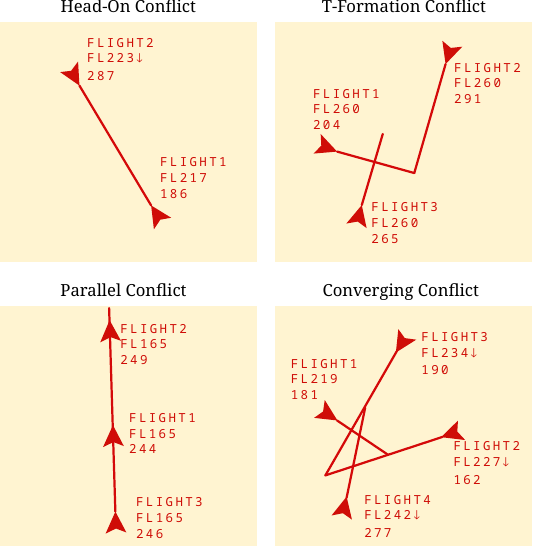}
    \caption{Examples of Conflict Scenarios}
    \label{fig:conflict_examples}
\end{figure}

In addition to conflict type, we also consider changes in flight levels. Some scenarios have all aircraft at the same level, while others involve climbing, descending, and level flights, adding further complexity to the conflict dynamics. Examples are shown in \autoref{fig:conflict_examples}.

All scenarios are designed under the assumption that, without timely intervention, the aircraft involved will inevitably collide. This design ensures that each scenario presents a genuine challenge that tests the models' abilities to effectively navigate and resolve potential airborne conflicts in high-risk situations.

It is also worth noting that all these scenarios present imminent conflicts with very short response time. They are incredibly challenging for human operators, especially when involving more than two aircraft.

\subsection{Results}

\begin{figure}[h]
    \centering
    \includegraphics[width=\columnwidth]{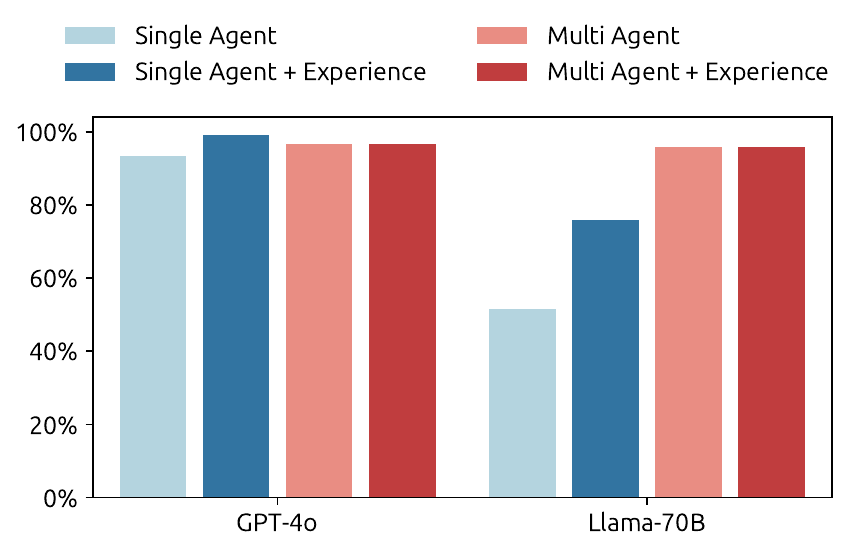}
    \caption{Success rate for different agent configurations, tested for a total of 120 conflict cases.}
    \label{fig:bar_chart_success_rate}
\end{figure}

These conflict scenarios are tested with single-agent and multiple-agent configurations using different language models.
\autoref{fig:bar_chart_success_rate} shows the success rates across different agent configurations for \texttt{GPT-4o} and \texttt{Llama3:70B} models.  We also test their performance when they have access to the \texttt{SearchExperimentLibrary()} tool.

For single-agent setup, we can see that \texttt{GPT-4o} performs better than \texttt{Llama3:70B}. And by including experience libraries, significant improvements are observed. For multiple-agent setup, the success rates are all high, even for the open-source \texttt{Llama3:70B} with a significantly smaller model size.

\begin{table}[ht]
\centering
\caption{Performance with single and multiple agents}
\label{tab:agent_config_results}
\footnotesize % Reducing the font size slightly more than \small
\begin{tabularx}{\columnwidth}{lXccc}
\toprule
\textbf{Model} & \textbf{Configuration} & \textbf{Collision} & \textbf{LoS} & \textbf{Resolved} \\ 
\midrule
\multirow{4}{*}{GPT-4o}
    & Single Agent & 4 & 4 & 112 \\
    & Single Agent + Exp & 0 & 1 & 119 \\
    & Multiple Agent  & 4 & 0 & 116 \\ 
    & Multiple Agent + Exp & 4 & 0 & 116 \\
\midrule
\multirow{4}{*}{Llama3:70B} 
    & Single Agent & 13 & 45 & 62 \\
    & Single Agent + Exp & 6 & 23 & 91 \\
    & Multiple Agent & 2 & 3 & 115 \\ 
    & Multiple Agent + Exp & 2 & 3 & 115 \\
\bottomrule
\end{tabularx}
\end{table}

\autoref{tab:agent_config_results} shows the exact number of times the conflicts resulted in the collision, loss of separation (LoS), and conflict resolved. We observe that the best result is achieved by the single-agent backed by \texttt{GPT-4o} with experience library, where only 1 out 120 was not fully cleared.

\begin{figure}[ht!]
    \centering
    \includegraphics[width=\columnwidth]{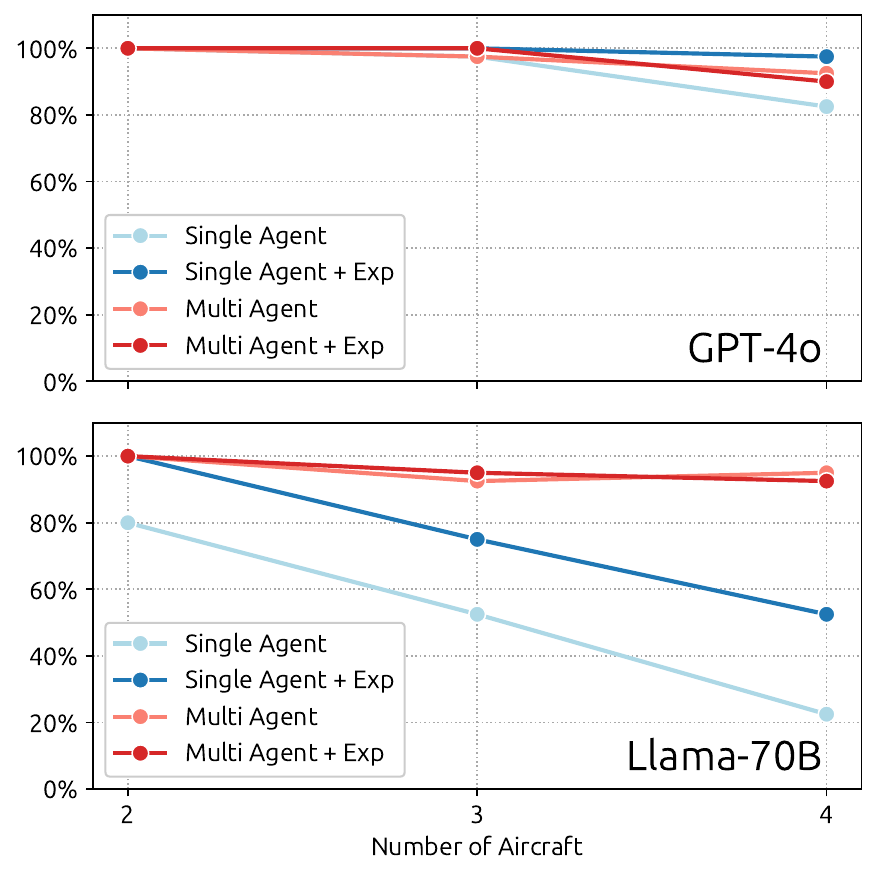}
    \caption{Success rate by the number of aircraft in conflict for different agent configurations, for a total of 120 cases.}
    \label{fig:success_rate_vs_num_ac}
\end{figure}

In \autoref{fig:success_rate_vs_num_ac}, we illustrate how the success rate of conflict resolution varies with the number of aircraft involved for both models. Here, we can observe the \texttt{GPT-4o} model manages to solve all conflicts for two-aircraft and three-aircraft cases and missed a few four-aircraft scenarios. Model \texttt{Llama3:70B} missed a few three-aircraft and four-aircraft cases in a multiple-agent setup.

% \hl{This relatively small dataset was primarily dictated by the current testing capabilities and resource constraints. Specifically, all models except \texttt{GPT-4o} are hosted on a cloud platform called Groq, which imposes token-per-minute restrictions. This limitation prevents running multiple models in parallel, necessitating a sequential testing approach and thereby reducing the number of scenarios that can be tested within a reasonable timeframe.}

% \begin{figure}[h]
%     \centering
%     \includegraphics[width=0.5\linewidth]{figures/heat_map_success_rate_model_vs_temp.png}
%     \caption{Heatmap of Success Rates by Model and Temperature from Initial Experiment}
%     \label{fig:heatmap_initial_results}
% \end{figure}
% \begin{figure}[h]
%     \centering
%     \includegraphics[width=0.5\linewidth]{figures/total_score_vs_model_best_tempt.png}
%     \caption{Caption}
%     \label{fig:enter-label}
% \end{figure}

% \hl{Describe the list of experiments with different scenarios.}

\section{Discussions} \label{sec:discussions}

\subsection{General observation on performance} 

The single-agent setup with \texttt{Llama3:70B} model demonstrated the weakest performance. This outcome can be related to the smaller model size and context window compared to \texttt{GPT-4o}. However, its performance can be significantly improved when utilizing the experience library, with its success rate improved from 52\% to 76\%. This suggests that an agent with a smaller LLM can make use of knowledge from the experience library to significantly enhance its problem-solving efficiency. 

Moreover, the smaller \texttt{Llama3:70B} model achieved the most optimal performance in a multi-agent configuration. In this setup, the distributed processing load allows each agent to handle less information. This is especially beneficial given the smaller context window of only 8,000 tokens available to \texttt{Llama3:70B}, allowing more efficient information processing and decision-making across multiple agents.

\texttt{GPT-4o} had a high success rate across all the agent configurations. The single-agent configuration without experience can already achieve a success rate of 93\%.  With the experience library, it only had a single unresolved conflict, arriving at a 99\% success rate. Multiple-agent setups with and without experience demonstrated similarly high performances, suggesting that the LLM's size and larger context window play a more significant role than the model architecture.

\subsection{Optimizing the use of experience library}
Throughout the development of the experience library, we have been constantly adapting the experience library architecture. Initially, the library included complete experience documents, but it was discovered that by embedding only the conflict description, we can enhance search accuracy and relevance. This change ensured that search results were more applicable to current conflicts.

We noticed more performance issues with the \texttt{Llama3:70B} when summarizing the experience, particularly its tendency to produce inaccurate content in experience documents. Again, this is due to the smaller size of the model. To maintain a consistent quality of the library's content, we only retain experience documents created by \texttt{GPT-4o}.

Sometimes, the experience retrieval may encounter problems, especially with incorrect metadata input, e.g., the number of aircraft in conflict. For example, \texttt{Llama3:70B} often incorrectly searched for two-aircraft conflict solutions when dealing with four-aircraft scenarios. We can solve this issue by using better system prompts to refine the search.

\subsection{Limitations due to language model hallucination} 

There are several reasons why the aircraft conflict is not resolved, and each agent configuration has its own reasons. Starting with the single \texttt{Llama3:70B} agent, the main reason is the model's size, which affects reasoning capabilities. 

Many times, an agent sends an altitude command, which will not ensure enough vertical separation even though there are instructions for vertical separation in the system prompt. An example below shows how FLIGHT3 is instructed to descend to 22800 feet, which would reduce the vertical separation between the rest of the aircraft.

\begin{lstlisting}
Invoking: `CONTINUEMONITORING` with `{'duration': 10}`
Aircraft Pairs in Conflict and their TCPA (sec):
...
<EXTENSIVE CONFLICT INFORMATION>
...
Number of aircraft in conflict: 3
Aircraft Altitude Information:
FLIGHT1: Altitude 22500.0 ft -> 22500.0 ft (level)
FLIGHT2: Altitude 23322.38 ft -> 23000.0 ft (descending)
FLIGHT3: Altitude 23328.64 ft -> 23298.75 ft (descending)
Invoking: `SENDCOMMAND` with `{'command': 'ALT FLIGHT3 22800'}`
\end{lstlisting}

Although the experience library contributes significantly to reducing errors, it is particularly effective at the beginning of conflict resolution. Initially, the agent applies the library's suggestions directly to its current conflict. However, if conflicts between aircraft pairs persist, the agent's limitations in reasoning may become apparent again.
Another current limitation often observed in agents is their failure to anticipate secondary conflicts caused by resolving initial ones. 
In the example below, the planner directed FLIGHT2 and FLIGHT4 to climb to the same altitude, which will later create a secondary conflict.

\begin{lstlisting}
Planner Agent:
Invoking: `GETALLAIRCRAFTINFO` with `{'command': 'GETACIDS'}`
...
Invoking: `CONTINUEMONITORING` with `{'duration': 10}`
...
1. **FLIGHT2**: Climb to an altitude of 36200 ft. This will create vertical separation from FLIGHT1 and FLIGHT3, reducing the risk of collision.
2. **FLIGHT3**: Descend to an altitude of 32200 ft. This will provide vertical separation from FLIGHT1 and FLIGHT2, ensuring safety.
3. **FLIGHT4**: Climb to an altitude of 36200 ft. This will increase vertical separation from FLIGHT1, reducing the risk of collision and ensuring FLIGHT4 is at a different altitude than FLIGHT3, which is descending.
\end{lstlisting}

Finally, it is still challenging for agents to decide whether to continue monitoring the airspace for an extended period or to intervene immediately. These are all areas for future improvements.
For example, there are cases where the planner agent develops a viable conflict resolution plan, but the verifier agent opts to re-plan. If the verifier agent had monitored the situation for a longer duration, the conflict would have been resolved without the need for re-planning.

\subsection{Complexity of the traffic}

Looking at \autoref{fig:success_rate_vs_num_ac}, it is evident that as the number of aircraft involved in a conflict increases, the success rate for resolving these conflicts declines. This outcome is expected as the more significant number of aircraft introduces more information that the large language model must process, which in turn impacts its performance. 

Notably, the \texttt{GPT-4o} agent configurations maintain similar performance levels when dealing with conflicts involving two or three aircraft. There is a slight decrease in performance when the number of aircraft increases to four. For \texttt{Llama3:70B}, the performance is similar in multi-agent setups.

\subsection{Computing resource constraints}

Our testing capabilities were significantly influenced by accessing computing resources, particularly in the context of model hosting and processing power. 

All models, with the exception of GPT-4o, were hosted on a cloud platform called Groq. Groq utilizes a specialized processing unit known as the Language Processing Unit, which can deliver around 1000 tokens per second, making it an optimal choice for our needs. However, Groq also imposes token-per-minute restrictions. This restriction prevented us from running multiple models and conflict scenarios in parallel, thus reducing the number of scenarios we could test.

We have also tried to set our own \texttt{Llama3:70B} model with Ollama and using TU Delft DelftBlue cluster \cite{DHPC2024}, where NVIDIA A100 GPUs are available. However, the inference speed in the high-performance computing environment is too slow for our use cases.

% \textbf{Dynamic Agent Number:}
% As shown in \autoref{fig:success_rate_vs_num_ac}, the performance of agents decreases as the number of aircraft involved in a conflict increases. While increasing the number of agents to three improves performance, even this multi-agent system experiences a decline when faced with a higher number of aircraft. To address this issue, a dynamic multi-agent system could be employed, where the number of agents is adjusted based on the number of aircraft in conflict.

\section{Conclusion} \label{sec:conclusion}

This study explored the application of large language models as embodied agents in air traffic control scenarios, focusing on their ability to autonomously resolve conflicts. 

Our experiments with both open and closed-source models such as \texttt{Llama3:70B} and \texttt{GPT-4o} demonstrate the huge of large language models embodied agents in performing air traffic control tasks. This new approach could reduce the gap between artificial and human situational awareness. We have demonstrated that it provides human-like reasoning with timely control instructions or recommendations.

The findings highlight that larger models outperform smaller models in complex conflict resolution scenarios. The incorporation of an experience library further aids in boosting efficiency by providing access to past conflict resolution insights, which is particularly beneficial for smaller models like \texttt{Llama3:70B}. 

Moreover, the study has shown that multi-agent systems, where tasks are distributed among specialized agents, yield high success rates in resolving conflicts as well. This research paves the way for new research paths to apply language model-embodied agents in more complex tasks for air traffic management.

\bibliographystyle{ieeetr}
\bibliography{references}

\end{document}